\documentclass[manuscript,screen]{acmart}

\usepackage{comment}
\usepackage[utf8]{inputenc}
\usepackage[inline]{enumitem}

\usepackage{setspace}
\usepackage{multicol}

\AtBeginDocument{%
  \providecommand\BibTeX{{%
    \normalfont B\kern-0.5em{\scshape i\kern-0.25em b}\kern-0.8em\TeX}}}

\setcopyright{acmcopyright}
\copyrightyear{2022}
\acmYear{2022}
\acmDOI{10.1145/1122445.1122456}

\acmJournal{JACM}
\acmVolume{37}
\acmNumber{4}
\acmArticle{111}
\acmMonth{8}

\begin{document}


\title{Healthsheet: Development of a Transparency Artifact for Health Datasets}
\author{Negar Rostamzadeh}

    \affiliation{%
    \institution{Google Research}
    \country{Canada}}
  \email{nrostamzadeh@google.com}
  
  \author{Diana Mincu}
      \affiliation{%
    \institution{Google Research}
    \country{UK}}

  \author{Subhrajit Roy}
     \affiliation{%
    \institution{Google Research}
        \country{UK}}

  \author{Andrew Smart}
 \affiliation{%
    \institution{Google Research}
        \country{US}}

  \author{Lauren Wilcox}
 \affiliation{%
    \institution{Google Research}
        \country{US}}

  \author{Mahima Pushkarna}
 \affiliation{%
    \institution{Google Research}
        \country{Canada}}

  \author{Jessica Schrouff}
 \affiliation{%
    \institution{Google Research}
        \country{UK}}

  \author{Razvan Amironesei}
 \affiliation{%
    \institution{Google Research}
        \country{US}}

  \author{Nyalleng Moorosi}
 \affiliation{%
    \institution{Google Research}
        \country{Lesotho}}

  \author{Katherine Heller}
 \affiliation{%
    \institution{Google Research}
    \country{US}}



\begin{abstract}
  Machine learning (ML) approaches have demonstrated promising results in a wide range of healthcare applications. Data plays a crucial role in developing ML-based healthcare systems that directly affect people's lives. Many of the ethical issues surrounding the use of ML in healthcare stem from structural inequalities underlying the way we collect, use, and handle data. Developing guidelines to improve documentation practices regarding the creation, use, and maintenance of ML healthcare datasets is therefore of critical importance. In this work, we introduce \textit{Healthsheet}, a contextualized adaptation of the original datasheet questionnaire ~\cite{gebru2018datasheets} for health-specific applications. Through a series of semi-structured interviews, we adapt the datasheets for healthcare data documentation. As part of the Healthsheet development process and to understand the obstacles researchers face in creating datasheets, we worked with three publicly-available healthcare datasets as our case studies, each with different types of structured data: Electronic health Records (EHR), clinical trial study data, and smartphone-based performance outcome measures. Our findings from the interviewee study and case studies show 1) that datasheets should be contextualized for healthcare, 2) that despite incentives to adopt accountability practices such as datasheets, there is a lack of consistency in the broader use of these practices 3) how the ML for health community views datasheets and particularly \textit{Healthsheets} as diagnostic tool to surface the limitations and strength of datasets and 4) the relative importance of different fields in the datasheet to healthcare concerns. 
\end{abstract}



\maketitle
\section{Introduction}

The use of machine learning (ML) is rapidly expanding in healthcare as the amount of new data generated improves our capacity to effectively manage complex clinical and diagnostic information~\cite{matheny2020artificial,gulshan2016development,tomavsev2019clinically,perez2019large,wu2010prediction}. During the last decade, ML has played a central role in high-stakes heathcare problems such as precision medicine \cite{uddin2019artificial, piccialli2021precision}, survival analysis \cite{nagpal2021deep,lee2019dynamic}, disease diagnosis \cite{faust2018deep}, and continuous innovations in treatment plans. While ML approaches present opportunities to assist healthcare professionals, streamline the healthcare system, and potentially improve patient outcomes, they bring with them ethical concerns \cite{mazurowski2020artificial,banerjee2021reading,gichoya2021equity} that range from racial and gender disparities, accessibility of clinical studies, to subjectivity in healthcare practices and biases \cite{chen2020ethical}. 

Many ethical concerns in ML applications can be traced back to the development processes of the underlying datasets \cite{sambasivan2021everyone, prabhu2020large}. 
As Parasidi et al \cite{parasidis2019belmont} argue, societal fairness dictates that health data be used to advance the public good in ways that can avoid causing or exacerbating inequities. These societal goals are reflected in the many declarations of ethical principles from AI firms and research institutions \cite{mittelstadt2019principles}. Limited ethical and social structural deliberation during data collection can also exacerbate unfair racial biases in healthcare predictions \cite{obermeyer2019dissecting}. Therefore, caution and improved accountability in data collection, use and maintenance are critical in the adoption of ML in the health domain. 

As a step towards more accountable ML practices\cite{hutchinson2021towards}, several guidelines and frameworks have been proposed around data transparency, collection and use in ML \cite{gebru2018datasheets, geiger2020garbage, raji2020closing, paullada2020data,bender2018data,benjamin2019towards}. For example, ``Datasheets for datasets'' \cite{gebru2018datasheets} provided a critical new approach for the rigorous and reflective documentation of ML datasets. The fundamental concepts that underlie datasheets have been adopted in various forms across industry and academia, such as Artsheet \cite{srinivasan2021artsheets}, Data Cards \cite{pushkarnadata}, and FactSheets \cite{arnold2019factsheets}. However, the value of dataset documentation extends beyond the documentation artifact to the process of meeting documentation requirements; thereby increasing opportunities for greater integrity and accountability in data collection practices.

To begin addressing gaps in current practices, frameworks, and standards for the ethical collection of health data for ML, we introduce a contextualized type of datasheet that attends to the needs of healthcare data: \textit{Healthsheet}. The purpose of \textit{Healthsheet} is to contribute to the meaningful ethical review of healthcare data, in addition to existing data governance practices or legal requirements of healthcare data. Further, it aligns with recent initiatives in clinical trials data collection \cite{rivera2020guidelines, liu2020reporting}, and data-driven digital health technologies \cite{ibrahim2021health}.

According to \cite{parasidis2019belmont}, existing regulatory frameworks that could cover health data are limited in their applicability to the general use of health datasets for ML. For example, The Health Insurance Portability and Accountability Act (HIPAA) does not mandate ethics review for data collection and downstream use. HIPAA also places no limits on the use of de-identified data, regardless of who controls the information \cite{parasidis2019belmont}. As \cite{parasidis2019belmont} argue, the General Data Protection Regulation (GDPR) and the California Consumer Privacy Act focus on notification, consent, and deletion rights, but this doesn't necessarily address issues about the ethical collection, documentation, and use of data \cite{parasidis2019belmont}.

Besides these limitations in regulatory frameworks, we pose three main questions: \begin{enumerate*} \item \textit{could the current state of health data documentation practices introduce challenges and blockers in equitable health research and practice?}, \item \textit{could dataset consumers use the Healthsheet to assess the quality of datasets?} and finally \item \textit{what would be the incentives for health dataset curators to create the Healthsheet documentation?}. \end{enumerate*}

To answer these questions and contextualize the datasheet, we first conducted a participatory study by interviewing 21 experts (dataset creators and consumers), with a wide range of expertise and diverse backgrounds, in relation to healthcare data. We then tested our new proposed framework on 3 publicly available datasets, each composed of a different data modality to examine the practicality and blockers in creation of \textit{Healthsheets}. In section \ref{method}, we describe the \textit{Healthsheet} questionnaire development processes. In section \ref{transparency}, we the define transparency as a step towards accountability. Section \ref{dive} details the original datasheet questionnaire \cite{gebru2018datasheets}. Section 5 discusses our expert interview methodology and findings. In section 6, we discuss the necessity and limitations of the datasheet when employed in healthcare applications of ML. We then present \textit{Healthsheet} case studies in \ref{cases} and corresponding findings. Finally, we conclude by discussing our study takeaways, the broader impacts and limitations of this work.

\vspace{-.2cm}
\section{Healthsheet Development Methodology}\label{method}
Integrating feedback from a variety of stakeholders meets the needs of more communities \cite{martin2020participatory}, particularly in high-stake scenarios such as healthcare \cite{schurgin2021isolation}. For this work, we created a team with multi-disciplinary expertise ranging from Healthcare, ML for Health, ML Fairness, Applied Ethics and Human-Centered Design. 

This study is framed as follows:
\begin{enumerate}
    \item We ground our research by co-defining transparency artifacts, and discuss how to use them as a step towards a more accountable dataset development process.
    \item Following a systemic review of ML for health literature, we created an initial adaptation of the datasheet \cite{gebru2018datasheets}, the \textit{Primary Healthsheet}, which was introduced as stimuli to expert participants.
    \item Through a set of expert interviews spanning backgrounds related to ML, health and socio-technical research, we (i) identified documentation shortcomings in health datasets and the impact of these shortcomings on research advancement and equitable healthcare practices, (ii) improved upon and used the \textit{Healthsheet} as a diagnositc tool for health datasets assessment, (iii) discussed the necessity and limitations of the \textit{Primary Healthsheet} questionnaire, \textit{and} (iv) discussed the potential incentives needed by dataset curators and extractors to create a \textit{Healthsheet} for their datasets. 
    \item We then validated our proposed \textit{Healthsheet} using 3 publicly available datasets, MIMIC-III \cite{MIMIC_III}, MSOAC \cite{larocca2018msoac} and Floodlight \cite{baker2020digital}, taking note of the process, limitations, challenges, as well as gaps in existing documentation.
    \item Finally, We made sure the corpus represents our interdisciplinary perspective through deep qualitative engagement across many iterations. 
\end{enumerate}

\section{Co-defining Transparency in healthcare dataset creation}\label{transparency}
Algorithmic decision-making and the role of data has been central to the discourse on algorithmic transparency. There are many definitions and characteristics of transparency that seek to serve the common goals of accountability and trust. \cite{datta2016algorithmic} define transparency as explaining decisions made by algorithmic systems, which enables the identification of harms introduced by algorithmic decision making, holds the entities in the decision making chain accountable for such practices, and detects errors in input data that cause adverse decisions. In providing explanations for an adverse decision, specific guidance and factors can be provided to reverse such a decision. Annany and Crawford suggest using the limitations of transparency as conceptual tools that pave the way towards accountability \cite{ananny2018seeing}. Kizilcec's work observed a bell-shaped curve between transparency and trust in an experiment in which transparency through explanations was measured as a factor of perceptions of understanding, fairness, accuracy of process, and trust in actors \cite{kizilcec2016much}. 

For the purposes of this work, we describe five salient characteristics of transparency that the authors arrived at through generative participatory methods \cite{craftplaybook} \textit{prior} to engaging with study participants:
\begin{enumerate*}
\item \textbf{Culture and Aspiration:} Transparency is a property of the systemic and organizational ecosystem around datasets.
\item \textbf{Knowledge Sharing and Management:} Transparency enables the fluid movement and absorption of knowledge across upstream, downstream, past, present, and future stakeholders and across projects themselves. 
\item \textbf{Access Management:} Transparency cannot be conflated with access. For transparency, aspects of access include disclosure about rationales and processes, and ensuring some degree of control to prevent misuse or uncalibrated use.
\item \textbf{Beyond Data Creation:} Transparency is not a tail-end consideration occurring towards the end of the \textit{project's} life cycle, but rather across the \textit{dataset's} life cycle, from the stage where requirements are gathered to when it is actively used in experimentation, research, and production.
\item \textbf{Motivating Factors:} Transparency should capture and foster respect for both the provenance of data \textit{and} the motivations for collecting and applying the data to solve the problems it was intended to address.
\end{enumerate*}
Our proposed Healthsheet template seeks to operationalize these characteristics of transparency.
\section{Healthsheets - Datasheets for healthcare}\label{dive}
Datasheets for datasets \cite{gebru2018datasheets}, was proposed as a means of standard communication between dataset creators and dataset consumers, specifically in the context of ML practitioners. Although generalization is the primary goal of ML, we depart from general-purpose datasheets to address the unique context and requirements of healthcare. For example, in data from clinical studies, inclusion criteria can impact on '\textit{who is represented in datasets}', subsequently affecting the model being trained and tested on. 
\textit{Healthsheets} is meant to extend many of the benefits of datasheets, specifically to the health setting: dataset consumers can make more informed decisions about using the dataset for a specific task, while dataset creators also have a tool for self-reflection on the dataset they create in order to establish the intended use of their dataset, discover hidden characteristics of their data that can impact the outcome, and reduce the potential harms that use of the dataset can create. All of these benefits can eventually lead to the creation and promotion of safer, more reliable and equitable ML datasets in healthcare. 
We should first investigate the unique challenges that dataset consumers have for choosing and working with Healthcare datasets. How Healthsheets could be used as a diagnostic and auditing tool for a better understanding of datasets, their strength and limitations. Finally, while the datasheets for datasets paper \cite{gebru2018datasheets} has been widely cited, \footnote{ as of the day of writing, Datasheets for datasets has 574 citations}, all but a few actually create a datasheet, instead acknowledging the potential usefulness of datasheets. This shows that although the benefits of using datasheets are evident to the community, there are insufficient incentives for many dataset curators or creators to provide datasheets for their datasets. 

In cases where datasheets have been produced, creators have found the results to be extremely beneficial and analyses reported even more surprising. For example, Bandy et al \cite{bandy2021addressing} found several thousand duplicated books and a significant skew in genre representation when they created a datasheet for the BookCorpus dataset, which has been used to train large language models in industry. Several indings like this indicate that there is yet an unmet need to document and examine the myriad datasets which form the basis of ML, which are particularly critical in healthcare contexts. For example, a datasest with thousands of duplicated records could lead to inaccurate results that might impact patient safety. 

We find that current research practice in ML for healthcare often lead with model development, following which researchers search for datasets and use cases: ``\textit{If I have an ML claim that this method is good for something. I may use an off-the-shelf dataset... such that actually making progress on whatever classes are associated with these datasets, really translates to progress on some real-world problem.}''(P3). Although this might be commonplace in current ML model testing workflows, it is important that nuances of datasets be taken into consideration during model development for positive real-world impact in healthcare.

\section{Interview Procedure and Findings}
We conducted 21 semi-structured interviews with experts with a wide range of applied, industrial and academic expertise. Participants were recruited using snowball sampling \cite{goodman1961snowball}, in which an initial round of interview participants were recruited through emails targeting experts from clinical, legal, policy, privacy, bio-ethics, healthcare-related ML research, healthcare-related sociotechnical research, and applied healthcare ML engineering (see table 1 in Appendix B for detailed information on the expertise). Clinical experience included specializations in ophthalmology, medical imaging, nephrology, surgery and cancer; Engineering experience spanned research, production, program management, and product management in healthcare contexts. Additional experts were recruited through referrals from previous interviewees. Consent was obtained and stimuli material was provided to experts at least three days prior to interviews. More detailed discussions on the interviewee is provided in Appendix B.

\subsection{Challenges faced by dataset consumers}

In general, we found that the lack of centralized and comprehensive documentation exacerbates the challenges that experts face when selecting and using datasets. Of many reported challenges, a majority of participants considered the following to be the most concerning when it comes to impact on outcomes: \textbf{availability and clarity of meta-data}, \textbf{labeling and subjectivity} in labeling, and \textbf{inclusion/ exclusion criteria}.

\par \textbf{Meta-data clarity and availability}: The most common issues cited during the interviews was limited or lacking centralized meta-data information. This could lead to additional difficulties and challenges in using datasets:\\
``\textit{How much missingness in data exists... if datasets are already made, ... clarity around whether it's on GitHub, whether you have to contact this particular board, just understanding that saves a lot of time and a lot of research resource... There's a mismatch between a clinical academic understanding in what and how clinical question can be operationalized into processes in terms of where's the data from devices captured, labeling subjectivity,...}''- P19, \textit{``it's difficult to understand what can be done and what cannot be done with the data … that is under a lot of variables that we need to take into account to answer that question … and sometimes what we have found is that the knowledge is not centralized in one place''}-P5. Beyond identifying the appropriate u of datasets, interview participants raised questions about the inherent credibility and utility:
\textit{``A challenge we start with is, which entities should we get in contact with, what kind of content was acquired … it just requires a lot of back and forth with the data provider to understand this. When they say sex do they mean sex at birth or self-identified sex like gender, what is it?... ''}- P4. \textit{``A good chunk of our datasets is more (of a) by-product … but the metadata was never applied with the intent to research on, but it was more applied with the intent to have a proper understanding of the patient, and the usual ICD codes''}- P8.
\par \textbf{Subjectivity and labeling} was stated as one of the most critical concerns that could contribute to disparity of outcomes. In many cases, there are disagreements even on definitions of gold standards:
\textit{``understanding how the labels are generated is often the most difficult aspect … It's a big issue with chest x-rays actually''}- P3. \textit{``So, for example, how is a heart attack defined? Is it by blood test? Is it by symptoms? Because those huge amounts of subjectivity and inclusion criteria in that, excludes women and ethnic minorities.''}- P2.\\
Related themes that were frequently raised during the interviews by ML researchers and clincians pertained to dataset versioning, specific changes that were made, the impact of changes, intentions for the change, labeling, and availability of labeling guidelines: \textit{``The other aspect, that you touched on in the questionnaire on labeling is, it's well understood that agreement between doctors on a given task is way below $100\%$ in which level of granularity would a human specialist go''}- P19.\\ 

\par \textbf{Inclusion criteria and accessibility} was the third most cited theme, often described as a leading cause of disparity of outcomes in clinical studies – a sentiment primarily expressed by clinicians. Geographic location of collected dataset, referring hospitals, criteria to participate in trials, and guidelines are all of immense importance in determining participant eligibility and selection in clinical studies: \textit{`` Some hospitals require referrals, some are teaching hospitals, some are general hospitals … If someone said to me, these are all the x-rays from X’s neonatal unit. I know what that means, but most people don't''.}-P2.\\ For example, the EDSS score \cite{kurtzke1983rating}, which is the most commonly used metric for assessing disabilities in Multiple Sclerosis (MS), was defined for men hospitalized for MS in the United States Army during World War II \cite{kurtzke2015origin}. One participant, in particular pointed to the unique impact of inclusion and exclusion criteria on datasets: \textit{``what are the inclusion criteria? … very different from a dataset like imagenet which is to some extent curated''.} \textit{``... in healthcare, there might be some particular exclusion criteria like all the clinical data for over 18s … pretty blunt and pretty obvious …''} - P2 \\
As stated above, some of the challenges were specifically brought up and discussed by certain roles. For example, details around data composition, and versioning was primarily raised by ML Researchers, whereas issues related to calibration of the devices, inclusion/exclusion criteria, sites of data collections, were mainly discussed by clinicians. Subjectivity in labeling and demographic information were discussed across multiple roles.

\subsection{Incentives of dataset curators to fill the Healthsheet questionnaire}
Before discussing the benefits of healthsheets for dataset consumer, an important first step is to investigate potential incentives that can encourage dataset curators and extractors to create healthsheets for their datasets.  The subjective nature of incentives was a repeated themes mentioned by several, largely dominated by a discussion of the trade-offs between the perceived effort and time required in creating a healthsheet and its impact. \textit{``… if I’ve done a lot of hard work in releasing datasets, it's fair to ask me to create a datasheet. That's kind of meant to be a well-curated thing. That's … an artifact … with real Impact. … it makes sense in that context to have something relatively detailed. … this shouldn't be the most time-consuming and energy consuming piece of that project.’’} -P3 . P5 mentions \textit{``People spend a ton of time when they're launching an open source tool investing in the documentation, right? I spend a ton of time not because I care about documentation itself, but because I know that's best practice for launching a tool.’’}. Notably, one participant described long-term gains over time from reduced overhead, as a result of a one-time investment in the creation of healthsheets:  \textit{``that is going to be a peace of mind for them… So, if we don't have this data handy, like, in a single place, … these teams are going to be on the hook to answer the same questions over and over and over and it's more overhead for them.’’}. \\
Many participants also acknowledged \textit{``we don't have good incentives in place right now to give people credit for creating datasheets.’’}-P5. 
This acknowledgement is key for our purposes because it allows us to move past an analysis of datasheets (and by extension, the modes of creation of datasets), as solely reliant on the subjective intentions of stakeholders producing these artifacts. P4 says \textit{``Imagine I created a datasets that is very weird or very particular and I kind of want to hide that a bit or bury it ... You really don't want to document these things. You'd hope this doesn't happen but I can imagine it happening … so I suppose the motivation has to be persuading people to use the datasets. I could imagine somewhere down the line, they're being sort of standardized reporting and Regulation and it being a sort of thing you have to do.’’}. Along this line P16 says,
\textit{``More processes are harder for people to do given that their focus is on the research itself … So, either a carrot or stick to get someone to fill this out. Right? So it's either a standard, and if you don't use the standard across the board, they are unwilling to ingest your data … or something that is going to make them really excited to fill it out’’}.\\P19 touches on valuing the contributions differently to change the incentives of the community \textit{``Publications are valued more than data sharing and maybe that should change. Like, for example, when you get evaluated for promotion, how many datasets have you shared? Because that impacts so much more than just having one publication on a very limited topic. So I think revamping the way we evaluate people, might help with Incentivizing.}-P19. In sum, incentives should not be solely based on personal intentions or motivation alone. This should be grounded by both the way the community and institutions rewarding the creation of these artifacts, as well as standardized guidelines. 

\subsection{Healthsheet as a diagnostic tool}
After discussing the challenges faced by dataset consumers, we posed questions about the utility of \textit{Healthsheets} to tackle the specific aforementioned issues, as well as general issues concerning the use of health datasets. All participants stated that they would use \textit{Healthsheets} if datasets were accompanied by them. The most common utilities of \textit{Healthsheets} reported were \textbf{understanding nuances and limitations of datasets} prior to their use; and an expected increase in the \textbf{efficiency and ease of use of datasets} heralded by accessible, clear and transparent meta-data. Below are quotes from the interviewees around their motivations for using \textit{Healthsheets}.

\par \textbf{Understanding nuances and limitations of datasets: }
\textit{``I would see it as a guide for surfacing the domain specific questions or things that we should be mindful of''}-P1.\\
\textit{``Struggles with data could be technical or non-technical. The technical stuff is easy, the non-technical considerations is not... knowing the important caveats of the datasets ... You need actually documented somewhere.''}- P4.
\textit{``Understanding the nuances of the datasets before you spend a lot of time working with it..., for example, if there's a dataset that is incredibly clean, so all the photos in the dataset are from specific angles, always with the same camera always in the same position, that's fine if that's the task you’re training the model for. If you're training the model for consumers it's terrible''}- P4. P4 continues on the importance of transparency in data documentation by stating \textit{``It feels morally easy to say that all datasets should cover everyone in the world, in a fair and Equitable way, and there's some truth to that but there's also practical real-world issues to deal with, and it's okay if datasets are not fully representative as long as you know that from the beginning.''}- P4.

\par \textbf{Efficient and easier use of datasets: }
\textit{``some datasets may be created 30 years ago, ... science was quite different and the data is very different from how people nowadays think about things ... there's a paradigm shift in the healthcare data, just like the healthcare practice. Having better documentation ... is definitely super helpful ... We care a lot about long-term follow-up, 10 years, 15 years, 20 years … and it is non-trivial to find documentation ...it becomes extremely challenging. So if people have this level of details documented, like 30 years ago, I think that would have made our life a lot easier''}- P3.\\
\textit{``We need to understand the data, where it's been used, or what are the possible usages of the data in order to advise teams, and also for ourselves to be able to store it securely ... we want to take as much of that information out of a written doc and make it enforceable through infrastructure and so teams have to think less about data and can focus on their work. So the more information we have about datasets, definitely can help.''}- P11. \textit{``For working on a product launch it is hugely important where the data came from, how large it is, and how to work on it''}- P7.

\section{Datasheet for healthcare deep dive}
In this section, we dive deep into different groups of questions in the original datasheet \cite{gebru2018datasheets}, analyze the effectiveness and adequacy of the questionnaire sections for healthcare applications relying on the case studies, health literature and interviews. The original datasheet consists of questions grouped by stages of dataset creation, from motivation, to data collection and maintenance. We also address the gaps that led to the development of \textit{Healthsheet}\footnote{all questions in this section with dagger sign, represents questions from original datasheet paper \cite{gebru2018datasheets} or with small modification}.

\subsection{Motivation}
The motivation section centres around incentives for the creation of datasets. \\``\textit{What was the purpose of creating the dataset? Were there specific gaps to be filled?}''$\dagger$.\\ Some ML datasets are created to understand an ML research question~\cite{krizhevsky2009learning} or examine the power of a learning approach. Some are collected to introduce a new task, or application that can be addressed by ML. 
It is important to explicitly mention the purpose for the creation of datasets, which helps both dataset creators and consumers to make informed decisions. In ML for healthcare, data is often collected for purposes such as effectiveness and side effects of medications, diagnostic tools, and disease prognosis. 

Datasheets also address the implicit motivations and reasons coming from \textit{``funding resources or research institutes''} involved in ML healthcare research. Questions on funding could give a better understanding of underlying incentives for choosing the given research problem and dataset. One of the historical examples of this issue is the tobacco companies funding lung cancer research \cite{brandt2012inventing}. In addition, transparency of funding could shed a light on which healthcare problems are prioritized and why.

Studies~\cite{chen2020ethical} show, there is a disparity of funding rate on problems that impact lower income and disadvantaged groups compared to the general population that could stem from social injustice. For example, Sickle Cell disease impacts mainly the Black population while cystic fibrosis impacts mainly the white population. Cystic fibrosis receives significantly more funding and research resources, despite both being genetic disorders of similar severity \cite{chen2020ethical, farooq2018disparities, park2010ncaa}. Multiple studies also show that health research on topics related to women's health does not get sufficient funding and attention~\cite{chen2020ethical, chakradhar2018discovery,eisenberg2018epidemiology,pierson2019menstrual}. 

Studies suggest that disparities in research teams demographics and backgrounds could impact funding priorities and exacerbate existing socioeconomic, racial, and gender injustices \cite{chen2020ethical,vidyasagar2006global,pierson2019menstrual}. To address this issue, in the Primary Healthsheet questionnaire motivation section, we added a question on the demographic disparity of their research teams behind dataset creation that initiated generative discussions and comments were divided. Most participants were really eager to keep this question. 4 participants suggested to instead highlight and prioritize questions, which were on (i) demographic information in the dataset such as inclusion criteria in clinical studies and data collection, (ii) labeler guidelines and subjectivity in labeling, and (iii) expertise of researchers involved in data collection. From these 4 participants, 2 were in favour of removing or modifying the question.\\
``\textit{I think the demographics of researchers can start to feel quite uncomfortable quite quickly. The classic example is Sickle Cell, which is massively underfunded. I think, though, I would push back on that because I would say, Let's say there's no sickle-cell dataset and there should be a better signal so we can explain that for all kinds of socio-cultural ways. But that doesn't actually change. The fact that what you've got now is a cystic fibrosis dateset, right? So if the entire research team was black, would it be still a cystic fibrosis research datasets or it wouldn't be.}''-P4.\\
``\textit{If you condition on the dataset itself, the demographic becomes an explanation as to why the data was created, but it adds very little to the actual interpretation of the datasets. And I think one thing that I would, I would be much more interested in the labels and subjectivity. So, for example, how is a heart attack defined? Is it by blood test? Is it by symptoms? Because those huge amounts of subjectivity and inclusion criteria in that, that excludes women and ethnic minorities.}''-P4.\\
One of the participants also mentioned the subjectivity in the definition of demographic disparities in teams.\\
``\textit{It's tricky because it makes the question quite subjective ... a research team of 10 white men from the same University and similar ages may say they working with two people who are not the same as them. They might think: oh well compared to the rest of the field, we have this wonderfully diverse team. I wouldn't like to blame them for that. But that's also missing important information}''-P2.\\ 
Multiple participants also mentioned the importance of highlighting the expertise of curators of datasets.\\
``\textit{if you're collecting a dermatological dataset, we'd want to make sure that the composition of the team includes clinicians, and, you know, it's not just ml researchers...}''-P6.\\
On the other hand, most participants were in favor of keeping the question and spoke out about the implications of this. One participant who was also a dataset creator mentioned the correlation that they observed between the gender diversity of the research team and availability of datasets. They also observed penitential relationships to other demographic attributes of researchers and dataset availability to public.\\
``\textit{We have a lot of projects, that I was hoping to convince people to think about this question. So one of them which is partially related is, we're looking at the researchers and publicly available datasets. So we systematically searched ML projects in Healthcare in 2019, ... we show nicely that there is diversity of researchers, if a dataset is publicly available ... So we've also improved the representation of minority researchers. If a dataset is publicly available versus if this is only a accessible to the internal investigators, tend to preserve the order of old white male, professors publishing off those datasets. So I think this is the evidence that we've been wanting like we everyone's talking about open science, but everything is a bit more theoretical when it comes to benefit.}''-P21.\\ After analyzing the data on this question, we add the following variation of the questions, that both touches on the demographic disparity and expertise of researchers:\\
\textit{What is the demographic population of dataset curators/ generators?} and \textit{What is the distribution of backgrounds and experience/expertise of the dataset curators/generators?}\\
In addition, we cover Health-related transparency questions related to the inclusion criteria, subjectivity of labeling and demographic information of people in datasets.  

Many Health datasets are released without predefined tasks and labels, or will be later labeled by researchers for specific tasks and problems. In addition, datasets in healthcare usually gets expanded over time. This dynamic updates of datasets for various reasons were brought up with all interviewees who worked directly with Health related datasets. 
It is important to report the rationals, motivations, and funding sources for initial data collection process as well as new versioning, tasks, and research problems. This aspect is thoroughly discussed in~\ref{versioning} and additional questions are suggested.

\subsection{Composition}
The composition section investigates the information that helps dataset consumers to make informed decision prior to using the dataset. 
Questions on the type and characteristics of the data in the dataset, target labeling, data splits, and all compositional aspects of the data are addressed in this section. Having a centralized documentation resource on composition of data was brought up mainly by ML researchers and engineers working with healthcare datasets or working on health-data related infrastructure. This was specifically highlighted as a time-efficient practice for both dataset creators and consumers. 
``\textit{there may be some information in data composition, for example, ... I received, like, ten emails a day asking me where did you store this information or how I can extract this tables from your data? even from a very superficial level of just like data acquisition, I will save time for myself...(by centralizing documentations).}''-P4
\par
``\textit{What do the instances that comprise the dataset represent?}$\dagger$'', and ``\textit{What data does each instance consist of?}''$\dagger$.\\ In healthcare scenarios, data is typically associated with people who are patients or subjects of the experiments.
Each subject/patient is associated with a hierarchy of information, which can span multiple time periods. For instance, a patient can have multiple admissions to the hospital, with each admission lasting multiple days. Similarly, a patient might be linked to multiple medical imaging cases, each including multiple images or different types of images (e.g., structural MRI and CT). This hierarchy of the information is an important aspect of the healthcare application, and is currently not captured in datasheets.

\par
``\textit{Is there a label or target associated with each instance?, Are there recommended data splits?}''$\dagger$

Target labels of datasets can be used to define a task based on the given data. In many healthcare scenarios, data is collected as a form of database, with data being associated with patients characteristics but not a specific task and consequently no data splits associated with it. Datasets in healthcare are often collected over the span of multiple years and as a result, multiple tasks could be associated with the data and problem definitions that were not of interest when the data collection process was started, may now be of importance. However due to the sensitivity and importance of healthcare applications, it is crucial to update the healthsheet for new tasks and applications that become associated with the dataset.
\par

``\textit{Does the dataset contain data that might be considered confidential?}''$\dagger$\\
For many Public Health datasets, it is important to ensure the confidentiality of data and data should be de-identified. In addition, compliance with data protection rules such as EU’s General Data Protection Regulation (GDPR), or comparable regulations in other jurisdictions such as Health Insurance Portability and Accountability Act (HIPAA) depending on the country of data collection and use, is of immense importance in healthcare applications. Data de-identification in healthcare should follow country-based rules and regulations.

\par``\textit{Is any information missing from individual instances?}''$\dagger$\\
In healthcare datasets, presence of censoring where the outcome is only partially observable happens frequently. This could be due to multiple reasons, including loss, no follow-up, patients choosing to skip survey questions, and accessibility issues. It is specifically important to identify loss of information due to accessibility issues, and question these aspects during the data collection process. 
\par
``\textit{Does the dataset identify any subpopulations (e.g., by age, gender)?}''$\dagger$\\
This question is very important on healthcare applications but it should also be expanded on various aspects, incentives for collection and use of demographic information, the way this information is collected and employed, and compliance with countries regulation should be considered.
\subsection{Collection, uses, distributions, and maintenance}
Developmental stages of a dataset from collection to use, distributions and suitability of datasets are crucial in health-specific applications.
\par 
\textit{``Who was involved in the data collection process?''}$\dagger$. In healthcare applications, clinicians, specialists, and patients themselves could be involved in providing information as part of the data collection process. For example, providing MRI reports, answering surveys in app datasets, assessing disability scales (e.g: EDSS in MS), functionality tests and pain scales. In healthcare, in addition to acknowledging the expertise of people involved in the data collection process, it is crucial to note that collected data could be very subjective. Studies show that EDSS scores could be very subjective in estimating disability in MS. It is reported that in many cases pain assessments is biased towards undermining the experiences of Black patients \cite{hoffman2016racial}.
\par
\textit{``Over what time-frame was the data collected?''}$\dagger$. \textit{``Has the dataset been used for any tasks already?''}$\dagger$, 
\textit{``What (other) tasks could the dataset be used for?''}$\dagger$, \textit{``Are there tasks for which the dataset should not be used?}$\dagger$\\ Healthcare datasets are often collected over span of multiple years. This contributes to shift in the tasks, intended use, and new research questions derived from the datasets. 
For example, MIMIC-III was created over the span of 11 years. Multiple versions of datasets are being produced in the process. Floodlight dataset is constantly being updated by new data entries.
Although that it is crucial to enable the use of datasets for new research questions and develop updated versions of them, it is important to have a comprehensive track of development processes, targeted research questions, intended uses and rational behind them.

\vspace{-.3cm}
\section{Healthsheet: Augmenting the datasheets for healthcare}\label{augment}
In this section, relying on literature, interviews, and our case studies, we discuss the aforementioned aspects that should be expanded or discussed in health context. 
\vspace{-.3cm}
\subsection{Dataset versioning}\label{versioning}
The amount of available Healthcare data has seen significant growth in recent years \cite{bighopesdata2020}. 
Multiple healthcare datasets, specifically datasets created using mobile apps such as Floodlight, are dynamic, meaning a new version of the dataset could be available in a short span of time. For some datasets, multiple versions of the same dataset may be released.

So, it is important to keep track of the properties that evolve with time. In the same way coding documentation is a living artifact, we propose that \textit{Healthsheets} be treated in a similar manner. 

Some datasets in healthcare are released without labels and predefined tasks, or will be later labeled by researchers for specific tasks and problems, to form sub-versions of the dataset. The following set of questions clarifies the information about the current version of the dataset.
It is important to report the rationale on labeling the data in any of the versions and sub-versions that this datasheet addresses, funding resources and motivations behind each released version of the dataset. Versioning aspects and the importance of having access to detailed documentation of records for variant versioning was mentioned with multiple interviewees. 
``\textit{In many cases, dataset curators don't really intend to create from the outset a dynamically updating dataset, like they kind of release the datasets and then some time passes and they're like, oh, we have some more data now. So now we're going to release again. I think there's very few institutions that go from the outset with the intention of dynamically updating.
}''-P17

\begin{enumerate*}
    \item Does the dataset get released as static versions or is it dynamically updated? 
    \begin{enumerate*}
    \item If static, how many versions of the dataset exist? 
    \item If dynamic, how frequently is the dataset updated?
    \end{enumerate*}
\item Is this datasheet created for the original version of the dataset? If not, which version of the dataset is this datasheet for?
\item  Are there any datasheets/ Healthsheets created for any versions of this dataset?
	\item  Does the current version/sub-version of the dataset come with predefined task(s), labels, recommended data splits (e.g, training, development/validation, testing)? If yes, please provide a high-level description of the introduced tasks, data splits, and labeling, and explain the rationale behind them. Please provide the related links and references. If not, is there any resource (website, portal, etc.) to keep track of all defined tasks and/or label definitions?
	\item If the dataset has multiple versions, and this datasheet represents one of them, answer the following questions:
	\begin{enumerate*}
	\item What are the characteristics that have been changed between different versions of the dataset?
	\item Explain the motivation/rationale for creating the current version of the dataset
	\item Does this version have more subjects/patients participating?
	\item Does this version of the dataset have extended data from the same patients of the older versions?
	\item Do we expect more versions of the dataset to be released?
	\item Is this datasheet for a sub-version of the dataset? If yes, does this sub-version of the dataset introduce a new task, labeling and recommended data splits? If the answer to any of these questions is yes, explain the rationale behind it.
	\item Are you aware of any widespread sub-version(s) of the dataset? If yes, what is the addressed task, or application that is addressed?
	\end{enumerate*}
\end{enumerate*}

As we move into the world of mobile health \cite{Steinhubl2015TheEF}, dataset creation will move from snapshots to a more fluid form where data points are continuously added. It is important to capture this distinction since it will inform the way a dataset is utilized.
 
In the case of static datasets, each sampling is done for a purpose. That motivation will surface in the dataset's properties, how many instances of each data point exist, where they were collected from, during what period. These properties change for each released version of a static dataset and should evolve with the versions.

\subsection{Accessibility in data collection}
One of the most cited aspects that was addressed during the interview process, was inclusion criteria. It is important to know who is involved in this dataset, and how. As P4 said: 
\textit{``What were the inclusion criteria? ... If it was only done in, let's say US hospitals, it's a limitation. If you could certainly apply to other US, hospitals, could you apply it to a Canadian or UK hospital? Possibly! Could you apply it to a South African Hospital? Maybe! Could you apply it to an Eritrean Hospital? Probably not. So, just understanding that kind of really headline stuff about recruitment where the data was from. It's going to depend on your deployment target, and I think it's very easy to say that.''}\\
In addition, many healthcare data collection systems exclude some patients/subjects due to the inaccessibility of data collection frameworks. Two of the datasets that we worked with are on Multiple Sclerosis patients, i.e: MSOAC \cite{larocca2018msoac} and Floodlight. Some patients with MS have movement issues such as tremor and this may make it harder for them interact with mobile apps. Some also have visual impairments. In order to have better patient representation, and consequently useful benchmarks, it is important to assess accessibility during the data collection process. In addition, most of the data collection processes only cover English language, which is not the default language for many patients. It is very important to be transparent about the limitation of the datasets with regarding to the accessibility aspects.
\begin{enumerate*}
\item Is there any language-based communications with patients? If yes, describe the choices of language(s) for communication. (e.g, if there is an app for communication, what are the language options in the app? (e.g: English, Spanish))
\item What are the accessibility measurements and what aspects were considered when the study was designed and implemented? (e.g, in a clinical study if it is accessible to someone with motor function disability? or if there is an app for patients with MS, are all aspects of inaccessibility associated with their potential disabilities considered?)
\item What are the inclusion criteria in the study? What is the recruitment strategy? (e.g., men over 18 years old, patients who are refereed from the hospital X)
\end{enumerate*}

\subsection{Collection and use of demographic information}

Racism and social conditions have had a fundamental impact on causes and prognosis of many diseases throughout the history~\cite{link1995social, wade2007ethics}. This has been increasingly evident during the COVID-19 pandemic. Data collected shows that BIPOC populations, specifically, Indigenous, Pacific islanders and Black populations have experienced the highest death toll from COVID-19\footnote{https://www.apmresearchlab.org/covid/deaths-by-race}~\cite{escobar2021racial}. On the one hand, given that socio-cultural factors can significantly affect the healthcare conditions of BIPOC populations, collection and access to demographic information is critical to conducting fairness analyses of ML healthcare models. 

On the other hand, modern American medicine has historical roots in scientific racism and eugenics movements, and racialized conceptions of susceptibility to disease persist to this day \cite{doi:10.1056/NEJMms2025396}. At the structural level, actions by parties ranging from medical schools to providers, insurers, health systems, legislators, and employers have ensured that racially segregated Black communities have limited and substandard care \cite{bailey2017structural}. Within medicine, there is no uniform practice regarding the use of race as a study variable and little to no expectation that authors examine racism as a cause of residual health inequities among racial groups \cite{boyd2020racism}. Likewise for ML, there is no uniform practice regarding the use of race as a variable in predictive models or datasets, nor is there an expectation that ML researchers examine structural racism as a cause of health inequities among racialized groups. 

Due to these historical and structural reasons, great care must be taken when collecting and using demographic information and specifically race in healthcare datasets. As Roberts \cite{roberts2018most} points out, "Using biological terms to define social inequities makes them seem natural—the result of inherent racial differences that can’t
be changed instead of unjust societal structures that must be dismantled." 

\cite{flanagin2021updated} recently released updated guidance on reporting of race and ethnicity in medical journals. These suggestions encourage fairenss, equity, consistency and clarity in the use and reporting of race and ethnicity in medical and science journals. The guidance explicitly recognizes race and ethnicity as social constructs, and we agree that this understanding is critical. In \textit{healthsheet}, we would like to identify whether the demographic information or variables for a dataset are collected and if yes/no, was there any rational or motivation behind it \footnote{Please note that the dept. of Health and Human Services (HHS) and FDA usually call out for demographic information to be collected. However, there is not a requirement of informed consent forms required by HHS. IRBs can require that information if they deem that information to be important to an individual's choice as to whether they participate in research or not}.

Does the dataset identify any demographic sub-populations (e.g., by age, gender, sex, ethnicity)? If yes,
\begin{enumerate*}
\item 
The reasons that these categories were assessed also should be described in the datasheet \cite{flanagin2021updated}.
\item Please describe who identified these categories and the source of the classifications used (e.g:  self-report or selection, investigator observed, database, electronic health record, survey instrument) \cite{flanagin2021updated}.
\item Are patients aware / consent to the collection and use of their demographic information?
\item Is there any task that any of these demographic information (e.g: gender, age, ethnicity) has any biologically proven impact on the outcome? If yes, please provide a link to the study, or publication.
\item Is there any mechanism for updating some of this information through the data collection process or afterwards? For example, if someone wants to change their gender information.
Provide a description of the respective distributions of each subgroup populations within the dataset. Is there any disparity of results between different demographics?
 \end{enumerate*}
If no, \begin{enumerate*}
\item Is there any regulation that prevents some of the gender / sex data collection in your study (for example, the country that the data is collected in)? 
\item Are you employing methods to reduce the disparity of error rate between different demographic subgroups when demographic labels are unavailable? Please describe.
\end{enumerate*}
\vspace{-.35cm}
\subsection{Devices and Contextual attributes in Data Collection}
The modality of a dataset refers to the different types of data it contains. Depending on the domain, healthcare datasets are created using a variety of devices and equipment. These devices come in various forms, ranging from large machines located in the hospitals (Magnetic resonance imaging (MRI), Computer Tomography Scans, X-rays, etc.) to small-scale wearable devices (eg. smart health trackers, wearable blood pressure monitors, etc.). Modalities not only govern the fidelity but also the statistical properties of the dataset. Hence, it is important to capture the details about the modalities of the dataset.
For data that requires a device or equipment for collection or the context of the experiment, answer the following additional questions or provide relevant information based on the device or context that is used (for example): 
\begin{enumerate*}
\item If there was an MRI machine used, what is the MRI machine and model used?, 
\item If heart rate was measured what is the device for heart rate variation that is used?, 
\item If cortisol measurement is reported at multi site, provide details, 
\item If smartphones were used to collect the data, provide the names of models.
\end{enumerate*}

\vspace{-.35cm}
\subsection{Labeling and subjectivity in labeling}
In medical domains, researchers usually take a dataset and appropriate it for a defined task. researchers may have their own guidance. It is important to know what the incentive of the original creators was, if there was a guideline or there is a guideline for the current version or sub-versions of the dataset?
\begin{enumerate}
\item Is there an explicit label or target associated with each data instance? respond for both the preliminary dataset and the current version.
\begin{enumerate}
    \item If yes:
    \begin{enumerate*}
        \item What are the labels provided?
        \item Who performed the labeling? For example, was the labeling done by a clinician, ML researcher, university or hospital?
    \end{enumerate*}
    \item What labeling strategy was used?
\begin{enumerate*}
    \item Gold standard label available in the data (e.g. cancers validated by biopsies)
    \item Proxy label computed from available data:
    \begin{enumerate*}

    \item Which label definition was used? (e.g. Acute Kidney Injury has multiple definitions)
    \item Which tables and features were considered to compute the label?
    \end{enumerate*}
    \item Which proportion of the data has gold standard labels?
   \end{enumerate*}

    \item Human-labeled data
    \begin{enumerate*}

    \item How many labellers were considered?
    \item What is the demographic of the labellers? (countries of residence, of origin, number of years of experience, age, gender, race, ethnicity, …)
    \item What guidelines did they follow?
    \item How many labellers provide a label per instance?
    \item If multiple labellers per instance:
    \begin{enumerate*}
    
    \item What is the rater agreement? How was disagreement handled? 
    \item Are all labels provided, or summaries (e.g. maximum vote)?
    
    \end{enumerate*}
    \item Is there any subjective source of information that may lead to inconsistencies in the responses? (e.g: multiple people answering a survey having different interpretation of scales, multiple clinicians using scores, or notes)
    \item On average, how much time was required to annotate each instance?
    \item Were the raters compensated for their time? If so, by whom and what amount? What was the compensation strategy (e.g. fixed number of cases, compensated per hour, per cases per hour)?
    \end{enumerate*}
\end{enumerate}
\end{enumerate}
\begin{enumerate*}\addtocounter{enumi}{1}
\item What are the human level performances in the applications that the dataset is supposed to address?
\item Is the software used to preprocess/clean/label the instances available? If so, provide a link or other access point. 
\item Is there any guideline that the future researchers are recommended to follow when creating new labels / defining new tasks?
\item Are there recommended data splits (e.g., training, development/validation, testing)? Are there units of data to consider, whatever the task? If so, provide a description of these splits, explaining the rationale behind them. provide the answer for both the preliminary dataset and the current version or any sub-version that is widely used.

\end{enumerate*}

\vspace{-.4cm}
\subsection{Challenge tests and confounding factors}
When developing models for healthcare, it is crucial to understand what limitations they will have. Such limitations stem partly from the data distribution covered: the narrower the data distribution, the more limited the use cases. This problem does not pertain to healthcare and generalization is a field of research on its own (see \cite{Pan2010-tx,Weiss2016-ch,Farahani2020-dl,Wang2021-fz} for reviews). 

In this section, we suggest to explicitly report limitations in the data distribution that might prevent generalization : \begin{enumerate*}
\item Which factors in the data might limit the generalization of potentially derived models? Is this information available as auxiliary labels for challenge tests? For instance:
\begin{enumerate*}
\item Number and diversity of devices included in the dataset, 
\item Data recording specificities, e.g., the view for a chest x-ray image, 
\item Number and diversity of recording sites included in the dataset. 
\item Distribution shifts over time (e.g., \cite{nestor19})
\end{enumerate*}

\item What confounding factors might be present in the data?
\begin{enumerate*}
\item Interactions between demographic or historically marginalized groups and data recordings, e.g., were women patients recorded in one site, and men in another?
\item Interactions between the labels and data recordings, e.g., were healthy patients recorded on one device and diseased patients on another?
\end{enumerate*}
\end{enumerate*}

\section{Case studies and Findings}\label{cases}
For the purpose of refining our framework, we used 3 publicly available datasets: MIMIC-III \cite{MIMIC_III}, MSOAC \cite{larocca2018msoac} and Floodlight \cite{baker2020digital}. 
\par \textbf{Reasons for choices of the datasets}: MIMIC-III is a clinical dataset. The reason that we chose MIMIC-III \cite{MIMIC_III}, was due to the availablity of the dataset, wide range of tasks, and applications on the dataset, availability and depth of documentation and details available from the dataset. In addition, multiple publications, employed the dataset to suggest new sub-version, tasks and labels on the dataset \cite{roy2021multi}. We also studied MSOAC \cite{larocca2018msoac} and Floodlight \cite{baker2020digital}, both on Multiple Sclerosis disease and are respectively on Electronic Health Records (EHR), and smartphone-based performance outcome measures. Floodlight is one of the examples of a dataset that has dynamic updates based on user data, and we chose it to reflect on our versioning and accessiblity questions.

We started by answering the original datasheet questionnaire, and identify the missing aspects, to be integrated in our \textit{Primary Healthsheet}. Once we had the finalized questionnaire, we went back to answer all questions from the start and identified some areas where necessary information is not publicly available or easily found. While this is due to the fact that we are not owners or creators of the dataset, it does outline the need for dataset creators to make use of such frameworks as the one we propose.\\
\textbf{Funding information: }
while for MIMIC-III and MSOAC this information was easy to find either on their official website or through online presentations, for Floodlight we could only make an assumption that it was sponsored by Roche based on the dataset website. 
\\
\textbf{Dataset statistics: }
MIMIC-III was the only dataset in the study that voluntarily presented such information. We recognize this is a time consuming process, but we would like to urge anyone creating a new dataset to consider releasing such information. It can be used as a test for identifying gaps in the data.\\
\textbf{Data acquisition: }
This was one of the hardest questions to find information on. Healthcare data consist of numerous fields, each of which could be collected through various mechanisms and very often these mechanisms are not specified anywhere. Even in the case of MIMIC-III, one of the more detailed datasets we used, for some fields the mechanism of acquisition is unknown.\\
\textbf{Demographics of dataset creators: }
Many researchers focus on diversity in the dataset itself, but when it comes to curating the data we often find that we do not know who was involved or included.
\\
\textbf{Versioning}:
Coming up with a common nomenclature for versions was a difficult task. We needed a term that would cover both conventional datasets that are collected and released once and continuously updated datasets. Some datasets such as MIMIC-III become so widely used that some research problems define their own version of a dataset which gains independent popularity.
The original Healthsheet creators cannot be expected to maintain all information derived from their original work. For this reason we expect that once the base Healthsheet is completed, the community will step up and add Healthsheet information for any new widely used version.

Another observation from the case studies was how difficult it was to find the information needed for the various sections. This meant that one source of information was usually not enough, we had to dig through official webpages, GitHub repositories, arXiv papers, news articles and more to pull the information and be sure it is accurate. This further outlines the need of a unified place where all this information lives. Further to completing a Healthsheet, we also propose that any entered information is officially verified and confirmed by the dataset owners, creators and maintainers.

Finally the true motivations behind the creation of each dataset can only be filled in by the groups responsible for them. As users of the datasets we can only assume the intention or infer it from publicly available resources, but us answering these questions will not influence the use of the data or the means of acquisition.

\section{Discussion}\label{discussion}
The field is well aware that algorithm design is not enough--we need to find ``good" datasets with which to evaluate algorithms. Yet, we have not had a community-wide standard to define what a``high-quality'' or ``good'' health dataset is. Healthsheet aims to bridge this gap and ensure that: 

\begin{enumerate*}
\item Data used in ML for health experiments is representative of real applications, with authentic and representative patient data.
\item Biases in datasets are more easily detected and understood, due to better means of describing and treating data types, and mitigating faulty assumptions about data types that arise due to lack of documentation.
\item Benchmark datasets are selected based on completeness and alignment with documentation standards, to make reliable, ethical comparisons of ML performance results possible.
\item Healthsheet could be a tool for retrospective and historical dataset analysis.
\item Future auditing could be conducted more easily.
 \item Healthsheet serves as a community-based artifact: with opportunities for review and evolution to meet community needs.
\end{enumerate*}

\noindent \textbf{Limitations:} We only worked with publicly available datasets. Majority of datasets in healthcare are not publicly availble and this on its own could add additional concerns around transparency. 
Our methodology for defining new questions was grounded on the expertise of the team of authors, interview participants, and datasets we studied. 

We hope that our work will start a conversation in the community and will continue to tailor \textit{Healthsheet} to broader healthcare scenarios over time. 
In addition, we believe that involving more stakeholders, such as patients or patients-representative communities, could immensely improve the \textit{Healthsheet} questionnaire, as a transparency artifact.
Finally, we would like to emphasize here that transparency and this questionnaire should not be thought of as ends by themselves but defined as means toward responsible and equitable ends.

\begin{acks}
We would like to thank Dr Madeleine Elish and Dr Berk Ustun and Noah Broestl, for their contributions to the earlier version of the paper presented at ML4H, Dr Timnit Gebru, Dr Xiao Liu, Dr Viknesh Sounderajah, Dr Leo Anthony Celi, Dr Mark Diaz, Dr Danielle Belgrave and Gerardo Flores for their valuable feedback, and last but not the least, all the interviewees who participated in this study and contributed in the formation of this paper.
\end{acks}

\bibliographystyle{ACM-Reference-Format}
\bibliography{sample-base}

\appendix

\newpage
\section*{Appendix A: Healthsheet}
The provided answers here are for the MIMIC-III dataset, that was one of our case studies. Several questions in the paper are from original datasheets \cite{gebru2018datasheets} or an adaptation of datasheets questionnaire. The Healthsheets questionnaire is being updated by ongoing interviews and feedback.
\definecolor{darkblue}{RGB}{46,25, 110}

\newcommand{\dssectionheader}[1]{%
   \noindent\framebox[\columnwidth]{%
      {\fontfamily{phv}\selectfont \textbf{\textcolor{darkblue}{#1}}}
   }
}

\newcommand{\dsquestion}[1]{%
    {\noindent \fontfamily{phv}\selectfont \textcolor{darkblue}{\textbf{#1}}}
}

\newcommand{\dsquestionex}[2]{%
    {\noindent \fontfamily{phv}\selectfont \textcolor{darkblue}{\textbf{#1} #2}}
}

\newcommand{\dsanswer}[1]{%
   {\noindent #1 \medskip}
}

\begin{singlespace}
\begin{multicols}{2}

\dssectionheader{General Information}
If the answer to any of the questions in the questionnaire is N/A, please describe why the answer is N/A (e.g: data not being available)
\dsquestionex{provide a 2 sentence summary of this dataset.}{}

\dsanswer{MIMIC (Medical Information Mart for Intensive Care) is a large, freely-available database comprising deidentified health-related data from patients who were admitted to the critical care units of the Beth Israel Deaconess Medical Center.}

\dsquestionex{Has the dataset been audited before?}{If yes, by whom and what are the results?}

\dsanswer{N/A. Information could not be easily found.}

\dssectionheader{Dataset Versioning}

\noindent \textbf{Version}: A dataset will be considered to have a new version if there are major differences from a previous release. Some examples are a change in the number of patients/participants, or an increase in the data modalities covered.
\newline \newline
\noindent \textbf{Sub-version}: A sub-version tends to apply smaller scale changes to a given version. Some datasets in healthcare are released without labels and predefined tasks, or will be later labeled by researchers for specific tasks and problems, to form sub-versions of the dataset.
\newline \newline
\noindent The following set of questions clarifies the information about the current (latest) version of the dataset. It is important to report the rationale for labeling the data in any of the versions and sub-versions that this datasheet addresses, funding resources, and motivations behind each released version of the dataset.

\dsanswer{}

\dsquestionex{Does the dataset get released as static versions or is it dynamically updated?}{
\newline a. If static, how many versions of the dataset exist?
\newline b.If dynamic, how frequently is the dataset updated?}

\dsanswer{Static versions:
\begin{itemize}
    \item MIMIC-IV contains data from 2008-2019. The data was collected from Metavision bedside monitors.
    \item MIMIC-III contains data from 2001-2012. The data was collected from Metavision and CareVue bedside monitors.
    \item MIMIC-II contains data from 2001-2008. The data was collected from CareVue bedside monitors. MIMIC-II is no longer publicly available but the data can still be obtained from MIMIC-III by only including data from the CareVue monitors.
\end{itemize}
}

\dsquestionex{Is this datasheet created for the original version of the dataset?}{If not, which version of the dataset is this datasheet for?}

\dsanswer{This datasheet is created for version 3 (MIMIC-III). There is no healthseet for the original dataset version.}

\dsquestionex{Are there any datasheets created for any versions of this dataset?}{}

\dsanswer{No, this is the first datasheet being created for this dataset.}

\dsquestionex{Does the current version/sub-version of the dataset come with predefined task(s), labels, and recommended data splits (e.g., for training, development/validation, testing)?}{If yes, please provide a high-level description of the introduced tasks, data splits, and labeling, and explain the rationale behind them. Please provide the related links and references. If not, is there any resource (website, portal, etc.) to keep track of all defined tasks and/or label definitions?}

\dsanswer{The original dataset doesn’t come with predefined tasks, labels or recommended data splits. There is a community with ongoing contributions (pull requests) and most of the dataset labeling / task definitions fall under those categories. See \url{https://github.com/MIT-LCP/mimic-code/tree/main/mimic-iii/concepts.}}

\dsquestionex{If the dataset has multiple versions, and this datasheet represents one of them, answer the following questions:}{
\newline a. What are the characteristics that have been changed between different versions of the dataset?
\newline b. Explain the motivation/rationale for creating the current version of the dataset.
\newline c. Does this version have more subjects/patients represented in the data, or fewer?
\newline d. Does this version of the dataset have extended data or new data from the same patients as the older versions? Were any patients, data fields, or data points removed? If so, why?
\newline e. Do we expect more versions of the dataset to be released?
\newline f. Is this datasheet for a sub-version of the dataset? If yes, does this sub-version of the dataset introduce a new task, labeling, and/or recommended data splits? If the answer to any of these questions is yes, explain the rationale behind it.
\newline g. Are you aware of any widespread sub-version(s) of the dataset? If yes, what is the addressed task, or application that is addressed?}

\dsanswer{
a. New data is being added (collected between 2008-2012). In addition, many data elements have been regenerated from the raw data in a more robust manner to improve the quality of the underlying data. For a full list of changes, please refer to \url{https://mimic.mit.edu/iii/about/releasenotes/}
\newline b. MIMIC-III is an extension of MIMIC-II: it incorporates the data contained in MIMIC-II (collected between 2001 - 2008) and augments it based on the changes described above (and in more detail in the release notes).
\newline c. Yes, new patients have been added.
\newline d. Yes, corrections to fields and new data has been included. See release notes for a full list of changes/additions.
\newline e. MIMIC-IV has been released, which is an update to MIMIC-III. While we cannot say with certainty, we expect more versions to be released in the future.
\newline f. It is not, while there are multiple sub-versions of MIMIC-III, we are including information for the overall version since this was a major release.
\newline g. MIMIC extract is widely used by the community. More details can be found at: \url{https://github.com/MLforHealth/MIMIC_Extract}}

\dssectionheader{Motivation}

\noindent Reasons and motivations behind creating the dataset, including but not limited to funding interests.
\newline \newline
\noindent For any of the following questions, if a healthsheet has already been created for this dataset, then refer to those answers when filling in the below information.

\dsanswer{}

\dsquestionex{For what purpose was the dataset created?}{Was there a specific task in mind? Was there a specific gap that needed to be filled? Please provide a description.}

\dsanswer{The dataset was created for research and development in electronic health records.}

\dsquestionex{What are the applications that the dataset is meant to address? (e.g., administrative applications, software applications, research)}{}

\dsanswer{Broadly healthcare research. Specific tasks were not set up as part of the dataset release.}

\dsquestionex{Are there any types of usage or applications that are discouraged from using this dataset?}{}

\dsanswer{No commercialization. From the agreement: "The LICENSEE will use the data for the sole purpose of lawful use in scientific research and no other."
}

\dsquestion{Who created this dataset (e.g., which team, research group) and on behalf of which entity (e.g., company, institution, organization)?}\textsuperscript{$\dagger$}

\dsanswer{ MIMIC was created through the work of researchers at the MIT Laboratory for Computational Physiology and their collaborators. See: \url{https://mimic.mit.edu/iii/about/acknowledgments/.}}

\dsquestionex{Who funded the creation of the dataset?}{If there is an associated grant, please provide the name of the grantor and the grant name and number.}$\dagger$

\dsanswer{Dataset creation was supported by grants from the National Institute of Biomedical
Imaging and Bioengineering (NIBIB) of the National Institutes of Health (NIH) under award numbers R01-EB001659 (2003-2013) and R01-EB017205 (2014-2018).}

\dsquestionex{What is the distribution of backgrounds and experience/expertise of the dataset curators/generators?}{}

\dsanswer{N/A. This information could not be easily found.}

\dsquestion{Any other comments?}

\bigskip
\dssectionheader{Data Composition}

\noindent \textbf{Instances:} Refers to the unit of interest. The unit might be different in the healthsheet compared to the downstream use case: an instance might relate to a patient in the database, but will be used to provide predictions for specific events for that patient, treating each event as separate.

\dsanswer{}

\dsquestionex{What do the instances that comprise the dataset represent (e.g., documents, images, people, countries)?}{Are there multiple types of instances? Please provide a description.}$\dagger$

\dsanswer{MIMIC is a relational database containing tables of data relating to patients who stayed within the intensive care units at Beth Israel Deaconess Medical Center. A table is a data storage structure which is similar to a spreadsheet: each column contains consistent information (e.g., patient identifiers), and each row contains an instantiation of that information (e.g., a row could contain the integer 340 in the patient identifier column which would imply that the row’s patient identifier is 340). A list of all tables can be found here: \url{https://mimic.mit.edu/iii/mimictables/}}

\dsquestionex{How many instances are there in total (of each type, if appropriate)?}{(breakdown based on schema, provide data stats)?}

\dsanswer{See \url{https://mit-lcp.github.io/mimic-schema-spy/}.}

\dsquestionex{How many patients / subjects does this dataset represent?}{Answer this for both the preliminary dataset and the current version of the dataset.}

\dsanswer{There are 46520 patients in total in the MIMIC-III dataset.}

\dsquestionex{Does the dataset contain all possible instances or is it a sample (not necessarily random) of instances from a larger set?}{If the dataset is a sample, then what is the larger set? Is the sample representative of the larger set (e.g., geographic coverage)? If so, please describe how this representativeness was validated/verified. If it is not representative of the larger set, please describe why not (e.g., to cover a more diverse range of instances, because instances were withheld or unavailable). Answer this question for the preliminary version and the current version of the dataset in question.}$\dagger$

\dsanswer{MIMIC is a relational database containing tables of data relating to patients who stayed within the intensive care units (ICU) at Beth Israel Deaconess Medical Center between 2001 and 2012. The dataset is designed to be representative of electronic health records in ICUs but may not be representative of general electronic health records data such as in the non-ICU hospital wards.}

\dsquestionex{What data modality does each patient data consist of?}{If the data is hierarchical, provide the modality details for all levels (e.g: text, image, physiological signal). Break down in all levels and specify the modalities and devices.}

\dsanswer{The dataset contains different types of clinical data such as:
\begin{itemize}
    \item Time-stamped nurse-verified physiological measurements (for example, hourly documentation of heart rate, arterial blood pressure, or respiratory rate).
    \item Documented progress notes by care providers.
    \item Continuous intravenous drip medications and fluid balances.
\end{itemize}}

\dsquestionex{What data does each instance consist of? “Raw” data (e.g., unprocessed text or images) or features?}{In either case, please provide a description.}$\dagger$

\dsanswer{No images are linked in MIMIC-III. All data is in the form of structured data, where each field comes from an electronic health record, and therefore has been processed.}

\dsquestionex{Is any information missing from individual instances?}{If so, please provide a description, explaining why this information is missing (e.g., because it was unavailable). This does not include intentionally removed information, but might include, e.g., redacted text.}$\dagger$

\dsanswer{The de-identification process for structured data required the removal of all eighteen of the identifying data elements listed in HIPAA, including fields such as patient name, telephone number, address and exact dates. Protected health information was removed from free text fields, such as diagnostic reports and physician notes.

There are other sources of missingness coming from the sparseness of the data, which is the nature of EHR. Not all lab values will be present at all times for a given patient for example.}

\dsquestionex{Are relationships between individual instances made explicit (e.g., They are all part of the same clinical trial, or a patient has multiple hospital visits and each visit is one instance)?}{If so, please describe how these relationships are made explicit.}$\dagger$

\dsanswer{All of the subjects are patients who stayed within the intensive care units at Beth Israel Deaconess Medical Center between 2001 and 2012.}

\dsquestionex{Are there any errors, sources of noise, or redundancies in the dataset?}{If so, please provide a description. (e.g., losing data due to battery failure, or in survey data subjects skip the question, radiological sources of noise)}$\dagger$

\dsanswer{There are redundancies when it comes to laboratory values which are repeated in the CHARTEVENTS table. This occurs because it is desirable to display the laboratory values on the patient’s electronic chart, and so the values are copied from the database storing laboratory values to the database storing the CHARTEVENTS.}

\dsquestionex{Is the dataset self-contained, or does it link to or otherwise rely on external resources (e.g., websites, tweets, other datasets)?}{If it links to or relies on external resources:
a. Are there guarantees that they will exist, and remain constant, over time
\newline b. Are there official archival versions of the complete dataset (i.e., including the external resources as they existed at the time the dataset was created)
\newline c. Are there any restrictions (e.g., licenses, fees) associated with any of the external resources that might apply to a future user? Please provide descriptions of all external resources and any restrictions associated with them, as well as links or other access points, as appropriate.}$\dagger$

\dsanswer{The dataset is self-contained, and the only information gathered from external resources is the date of death. This is acquired from the social security death registry.}

\dsquestionex{Does the dataset contain data that might be considered confidential (e.g., data that is protected by legal privilege or by doctor-patient confidentiality, data that includes the content of individuals non-public communications)?}{If so, please provide a description.}$\dagger$

\dsanswer{N/A. We believe it does not, but could not guarantee.
}

\dsquestionex{Does the dataset contain data that, if viewed directly, might be offensive, insulting, threatening, or might otherwise cause anxiety?}{If so, please describe why.}$\dagger$

\dsanswer{We do not believe so.}

\dsquestionex{If the dataset has been de-identified, were any measures taken to avoid the re-identification of individuals?}{Examples of such measures: removing patients with rare pathologies or shifting time stamps.}$\dagger$

\dsanswer{The dataset has been de-identified and dates/times have been shifted.}

\dsquestionex{Does the dataset contain data that might be considered sensitive in any way (e.g., data that reveals racial or ethnic origins, sexual orientations, religious beliefs, political opinions or union memberships, or locations; financial or health data; biometric or genetic data; forms of government identification, such as social security numbers; criminal history)?}{If so, please provide a description.}$\dagger$

\dsanswer{It contains health data, but in a non-identifiable way. For each patient, the following fields are specified:

\begin{itemize}
    \item insurance
    \item language
    \item religion
    \item marital status
    \item ethnicity
    \item gender (genotypical sex)
\end{itemize}
}

\noindent \framebox[\columnwidth]{%
\textbf{Devices and contextual attributes in data collection}
}
\newline

\dsquestionex{For data that requires a device or equipment for collection or the context of the experiment, answer the following additional questions or provide relevant information based on the device or context that is used (for example)
}{
\newline a. If there was an MRI machine used, what is the MRI machine and model used?
\newline b. If heart rate was measured what is the device for heart rate variation that is used?
\newline c. If cortisol measurement is reported at multi site, provide details.
\newline d. If smartphones were used to collect the data, provide the names of models.
\newline e. Anything else?}

\dsanswer{N/A. We could not find information related to this.}

\noindent \framebox[\columnwidth]{%
\textbf{Challenge in tests and confounding factors}
}
\newline

\dsquestionex{Which factors in the data might limit the generalization of potentially derived models?}{Is this information available as auxiliary labels for challenge tests? For instance:
\newline a. Number and diversity of devices included in the dataset.
\newline b. Data recording specificities, e.g., the view for a chest x-ray image.
\newline c. Number and diversity of recording sites included in the dataset.
\newline d. Distribution shifts over time.}

\dsanswer{Beth Israel Deaconess Medical Center was the only site from which data was collected. It is based in Boston, MA, USA.

For distribution shifts, in 2008 the EHR system changed from CareVue to MetaVision. Data which could not be merged is given a suffix to denote the data source. There are also smaller shifts in non-transition years as the patient distribution is non-stationary.

For all other questions we could not find the information.}

\dsquestionex{What confounding factors might be present in the data?}{
a. Interactions between demographic or historically marginalized groups and data recordings, e.g., were women patients recorded in one site, and men in another?
\newline b. Interactions between the labels and data recordings, e.g., were healthy patients recorded on one device and diseased patients on another?}

\dsanswer{a. As there is a single recording site, all groups have had data recorded from the same place. However, we are aware of work that looks into site-wise demographic differences.
b.  N/A, as labels aren’t associated with this version.}

\dsquestion{Any other comments?}

\bigskip
\dssectionheader{Collection and use of demographic information}

\dsanswer{}

\dsquestionex{Does the dataset identify any demographic sub-populations (e.g., by age, gender, sex, ethnicity)?}{If yes:
\newline a. The reasons that these categories were assessed also should be described in the datasheet.
\newline b. How was this information acquired? Please describe who identified these categories and the source of the classifications used (e.g:  self-report or selection, investigator observed, database, electronic health record, survey instrument).
\newline c. If patients’ demographic data is included, are patients aware / did they consent to the collection and use of their demographic information?
\newline d. In some cases, there have been biologically proven associations between demographics and the outcome. Are you aware of similar associations in the tasks covered by this dataset? Should users be wary of specific proxies or associations when using the dataset? If yes, please provide a link to the study, or publication.
\newline e. Is there any mechanism for updating some of this demographic information after its initial collection? For example, if someone wants to change their gender information, what are the mechanisms to do so?
\newline f. Provide a description of the respective distributions of each subgroup population within the dataset.}

\dsanswer{Most information could not be easily found in official sources.

For f. Subgroup information:
\begin{itemize}
    \item Female: 44.1\%, Male 55.9\%
    \item Asian 2.3\%, Black 9.4\%, Hispanic 3.27\%, White 70.89\%, Other 14.04\%
\end{itemize}}

\dsquestionex{If no, is there any regulation that prevents demographic data collection in your study (for example, the country that the data is collected in)?}{}

\dsanswer{N/A, as the dataset contains demographic information.}

\bigskip
\dssectionheader{Pre-processing / de-identification}

\dsquestionex{Was there any pre-processing for the de-identification of the patients?}{Provide the answer for the preliminary and the current version of the dataset.}

\dsanswer{The data was de-identified in accordance with Health Insurance Portability and Accountability Act (HIPAA) standards using structured data cleansing and date shifting.}

\dsquestionex{Was there any pre-processing for cleaning the data?}{Provide the answer for the preliminary and the current version of the dataset}

\dsanswer{Various fields have been cleaned through harmonization or de-duplication. The following release notes provide ample information as to what was changed and how: \url{https://mimic.mit.edu/docs/iii/about/releasenotes/}. Dates have also been date shifted as part of the deidentification process.}

\dsquestionex{Was the “raw” data (post de-identification) saved in addition to the preprocessed/cleaned data (e.g., to support unanticipated future uses)?}{If so, please provide a link or other access point to the “raw” data.}

\dsanswer{N/A. This information could not be found.}

\dsquestionex{Were instances excluded from the dataset at the time of preprocessing?}{If so, why? For example, instances related to patients under 18 might be discarded.}

\dsanswer{In the MIMIC-IV cohort, patients who underwent extubation during ICU stays were included. The exclusion criteria were as follows: (i) age <18 years, (ii) unplanned extubation, (iii) not the first extubation during the hospital stay, or (iv) no MV records before extubation.

Reference: \url{https://www.frontiersin.org/articles/10.3389/fmed.2021.676343/full}}

\dsquestionex{If the dataset is a sample from a larger set, what was the sampling strategy (e.g., deterministic, probabilistic with specific sampling probabilities)? }{Answer this question for both the preliminary dataset and the current version of the dataset.}

\dsanswer{The data is sliced based on time and collected from medical records information systems. No specific sampling has been used.}

\dsquestion{Any other comments?}

\dsanswer{
}

\bigskip
\dssectionheader{Labeling and subjectivity of labeling}

\dsquestionex{Is there an explicit label or target associated with each data instance?}{Please respond for both the preliminary dataset and the current version.
\newline a. If yes:
\newline i) What are the labels provided?
\newline ii) Who performed the labeling? For example, was the labeling done by a clinician, ML researcher, university or hospital?
\newline b. What labeling strategy was used?
\newline i) Gold standard label available in the data (e.g., cancers validated by biopsies)
\newline ii) Proxy label computed from available data: \newline 1. Which label definition was used? (e.g., Acute Kidney Injury has multiple definitions)
\newline 2. Which tables and features were considered to compute the label?
\newline iii) Which proportion of the data has gold standard labels?
\newline c. Human-labeled data
\newline i) How many labellers were considered?
\newline ii) What is the demographic of the labellers? (countries of residence, of origin, number of years of experience, age, gender, race, ethnicity, …)
\newline iii) What guidelines did they follow?
\newline iv) How many labellers provide a label per instance?
If multiple labellers per instance:
\newline 1. What is the rater agreement? How was disagreement handled?
\newline 2. Are all labels provided, or summaries (e.g., maximum vote)?
\newline v) Is there any subjective source of information that may lead to inconsistencies in the responses? (e.g: multiple people answering a survey having different interpretation of scales, multiple clinicians using scores, or notes)
\newline vi) On average, how much time was required to annotate each instance?
\newline vii) Were the raters compensated for their time? If so, by whom and what amount? What was the compensation strategy (e.g., fixed number of cases, compensated per hour, per cases per hour)?
}

\dsanswer{}

\dsquestionex{What are the human level performances in the applications that the dataset is supposed to address?}{}

\dsanswer{}

\dsquestionex{Is the software used to preprocess/clean/label the instances available?}{If so, please provide a link or other access point.}

\dsanswer{}

\dsquestionex{Is there any guideline that the future researchers are recommended to follow when creating new labels / defining new tasks?}{}

\dsanswer{}

\dsquestionex{Are there recommended data splits (e.g., training, development/validation, testing)?}{Are there units of data to consider, whatever the task? If so, please provide a description of these splits, explaining the rationale behind them. Please provide the answer for both the preliminary dataset and the current version or any sub-version that is widely used.}

\dsanswer{}

\dsquestion{Any other comments?}

\dsanswer{No questions were answered in this section, as this version does not come with labels.}

\bigskip
\dssectionheader{Collection Process}

\dsquestionex{Were any REB/IRB approval (e.g., by an institutional review board or research ethics board) received?}{If so, please provide a description of these review processes, including the outcomes, as well as a link or other access point to any supporting documentation.}

\dsanswer{The project was approved by the Institutional Review Boards of Beth Israel Deaconess Medical Center (Boston, MA) and the Massachusetts Institute of Technology (Cambridge, MA).}

\dsquestionex{How was the data associated with each instance acquired?}{Was the data directly observable (e.g., medical images, labs or vitals), reported by subjects (e.g., survey responses, pain levels, itching/burning sensations), or indirectly inferred/derived from other data (e.g., part-of-speech tags, model-based guesses for age or language)? If data was reported by subjects or indirectly inferred/derived from other data, was the data validated/verified? If so, please describe how.}

\dsanswer{Data was collected in-hospital by clinical staff, collected in the critical care unit or from the hospital record system. For example, for labs, a member of the clinical staff acquires a fluid from a site in the patient’s body, labels it and sends it for processing.}

\dsquestionex{What mechanisms or procedures were used to collect the data (e.g., hardware apparatus or sensor, manual human curation, software program, software API)?}{How were these mechanisms or procedures validated? Provide the answer for all modalities and collected data. Has this information been changed through the process? If so, explain why.}

\dsanswer{Two different critical care information systems were used for collecting the dataset: Philips CareVue Clinical Information System (models M2331A and M1215A; Philips Health-care, Andover, MA) and iMDsoft MetaVision ICU (iMDsoft, Needham, MA).}

\dsquestion{Who was involved in the data collection process (e.g., patients, clinicians, doctors, ML researchers, hospital staff, vendors, etc) and how were they compensated (e.g., how much were contributors paid)?}

\dsanswer{We could only find high-level information regarding the groups involved.

MIMIC is made available largely through the work of researchers at the MIT Laboratory for Computational Physiology and the following research groups:
\begin{itemize}
    \item Beth Israel Deaconess Medical Center
    \item MIT Clinical Decision Making
    \item MIT Computational Physiology and Clinical Inference Group
    \item MIT Lab for Computational Physiology
    \item Philips Health Care
\end{itemize}
}

\dsquestionex{Over what timeframe was the data collected? Does this timeframe match the creation timeframe of the data associated with the instances (e.g., recent crawl of old news articles)?}{If not, please describe the timeframe in which the data associated with the instances was created.}

\dsanswer{The data was collected between 2001 and 2012, however the dates in the dataset have been time shifted in order to help with de-identification.}

\dsquestionex{Does the dataset relate to people?}{If not, you may skip the remaining questions in this section.}

\dsanswer{
}

\dsquestion{Did you collect the data from the individuals in question directly, or obtain it via third parties or other sources (e.g., hospitals, app company)?}

\dsanswer{Third parties or other sources. MIMIC-III consists of data collected from two different clinical information systems, CareVue and MetaVision.}

\dsquestionex{Were the individuals in question notified about the data collection?}{If so, please describe (or show with screenshots or other information) how notice was provided, and provide a link or other access point to, or otherwise reproduce, the exact language of the notification itself.}

\dsanswer{N/A. This information could not be found.}

\dsquestionex{Did the individuals in question consent to the collection and use of their data?}{If so, please describe (or show with screenshots or other information) how consent was requested and provided, and provide a link or other access point to, or otherwise reproduce, the exact language to which the individuals consented.}

\dsanswer{Requirement for individual patient consent was waived because the project did not impact clinical care and all protected health information was de-identified.}

\dsquestionex{If consent was obtained, were the consenting individuals provided with a mechanism to revoke their consent in the future or for certain uses?}{If so, please provide a description, as well as a link or other access point to the mechanism (if appropriate).}

\dsanswer{N/A, as consent was not obtained.}

\dsquestionex{In which countries was the data collected?
}{}

\dsanswer{The dataset was collected in Boston, Massachusetts, United States of America.}

\dsquestionex{Has an analysis of the potential impact of the dataset and its use on data subjects (e.g., a data protection impact analysis) been conducted?}{If so, please provide a description of this analysis, including the outcomes, as well as a link or other access point to any supporting documentation.}

\dsanswer{N/A. This information could not be found.}

\noindent \framebox[\columnwidth]{%
\textbf{Inclusion Criteria- Accessibility in Data Collection}
}
\newline

\dsquestionex{Is there any language-based communication with patients?}{If yes, describe the choices of language(s) for communication. (for example, if there is an app used for communication, what are the language options?)}

\dsanswer{To the best of our knowledge, verbal communication was used with the patients. We could not find information on the various language accommodations that were made.}

\dsquestionex{What are the accessibility measurements and what aspects were considered when the study was designed and implemented?}{}

\dsanswer{N/A. We could not find this information in official sources.}

\dsquestionex{If data is part of a clinical study, what are the inclusion criteria?}{}

\dsanswer{N/A. We could not find this information in official sources.}

\dsquestion{Any other comments?}

\dsanswer{}

\bigskip
\dssectionheader{Uses}

\dsquestionex{Has the dataset been used for any tasks already?}{If so, please provide a description.}

\dsanswer{Yes. A common example includes “Length of stay in the ICU”.}

\dsquestionex{Does using the dataset require the citation of the paper or any other forms of acknowledgement?}{If yes, is it easily accessible through google scholar or other repositories.}

\dsanswer{es, citations are required and the citation format depends on the version used. The following page provides the citation format: \url{https://mimic.mit.edu/docs/about/acknowledgments/} depending on the version.

The paper itself is also linked in the documentation website and easily accessible through Google Scholar or online sources.}

\dsquestionex{Is there a repository that links to any or all papers or systems that use the dataset?}{If so, please provide a link or other access point.}

\dsanswer{There is no official repository available, but one could find papers through looking up the citation (e.g., \url{https://read.qxmd.com/keyword/229497})}

\dsquestionex{Is there anything about the composition of the dataset or the way it was collected and preprocessed/cleaned/labeled that might impact future uses?}{For example, is there anything that a future user might need to know to avoid uses that could result in unfair treatment of individuals or groups (e.g., stereotyping, quality of service issues) or other undesirable harms (e.g., financial harms, legal risks) If so, please provide a description. Is there anything a future user could do to mitigate these undesirable harms?}

\dsanswer{We do not have enough information to accurately answer this question. However the methods through which various fields were collected could lead to a bias in results, especially if there is a difference in care for an individual or a group.}

\dsquestionex{Are there tasks for which the dataset should not be used?}{If so, please provide a description. (for example, dataset creators could recommend against using the dataset for considering immigration cases, as part of insurance policies)}

\dsanswer{There is no official banning of any specific task, however research that could lead to the enforcement of any rule or law should be validated through other means.}

\dsquestion{Any other comments?}

\bigskip
\dssectionheader{Dataset Distribution}

\dsquestionex{Will the dataset be distributed to third parties outside of the entity (e.g., company, institution, organization) on behalf of which the dataset was created?}{If so, please provide a description.}

\dsanswer{Yes, anyone can access the dataset provided they meet the following criteria:
\begin{itemize}
    \item Become a credentialed user on PhysioNet. This involves completion of a training course in human subjects research.
    \item Sign the data use agreement (DUA). Adherence to the terms of the DUA is paramount.
    \item Follow the tutorials for direct cloud access (recommended), or download the data locally.
\end{itemize}}

\dsquestionex{How will the dataset be distributed (e.g., tarball on website, API, GitHub)}{Does the dataset have a digital object identifier (DOI)?}

\dsanswer{The data can be accessed on Cloud through either BigQuery or AWS; Alternatively it can be downloaded locally from PhysioNet.}

\dsquestion{When will the dataset be distributed?}

\dsanswer{The dataset was released on the 2nd of September 2016. Anyone can download it after they are approved for access.}

\dsquestionex{Assuming the dataset is available, will it be/is the dataset distributed under a copyright or other intellectual property (IP) license, and/or under applicable terms of use (ToU)?}{If so, please describe this license and/or ToU, and provide a link or other access point to, or otherwise reproduce, any relevant licensing terms or ToU, as well as any fees associated with these restrictions.}

\dsanswer{Yes, there is a user agreement that a researcher must agree to before getting access. The terms and conditions can be found in the PhysioNet account required during setup, and need to be accepted for each version of the dataset that you intend to use.}

\dsquestionex{Have any third parties imposed IP-based or other restrictions on the data associated with the instances?}{If so, please describe these restrictions, and provide a link or other access point to, or otherwise reproduce, any relevant licensing terms, as well as any fees associated with these restrictions.}

\dsanswer{Any results obtained from the work on the dataset should be open sourced to benefit the community. There are no restrictions concerning who owns the IP.}

\dsquestionex{Do any export controls or other regulatory restrictions apply to the dataset or to individual instances?}{If so, please describe these restrictions, and provide a link or other access point to, or otherwise reproduce, any supporting documentation.}

\dsanswer{N/A. The information required to answer this question could not be easily found.}

\dsquestion{Any other comments?}

\dsanswer{
}

\bigskip
\dssectionheader{Maintenance}

\dsquestion{Who will be supporting/hosting/maintaining the dataset?}

\dsanswer{MIT Laboratory for Computational Physiology and their collaborators (\url{https://mimic.mit.edu/docs/about/acknowledgments/}).}

\dsquestion{How can the owner/curator/manager of the dataset be contacted (e.g., email address)?}

\dsanswer{It is generally recommended to raise any issues with MIMIC Code Repository GitHub. If there are issues related to sensitive information then one can contact phi-report@physionet.org (for PHI) and mimic-support@physionet.org (for general private issues)}

\dsquestionex{Is there an erratum?}{If so, please provide a link or other access point.}

\dsanswer{ No erratum as of now, but release notes with corrections are published periodically (\url{https://mimic.mit.edu/docs/iii/about/releasenotes/}).}

\dsquestionex{Will the dataset be updated (e.g., to correct labeling errors, add new instances, delete instances)?}{If so, please describe how often, by whom, and how updates will be communicated to users (e.g., mailing list, GitHub)?}

\dsanswer{Yes. The maintainers of the dataset provide regular updates. These are communicated through the release notes page.}

\dsquestionex{If the dataset relates to people, are there applicable limits on the retention of the data associated with the instances (e.g., were individuals in question told that their data would be retained for a fixed period of time and then deleted)?}{If so, please describe these limits and explain how they will be enforced.}

\dsanswer{N/A. We could not find information about this topic.}

\dsquestionex{Will older versions of the dataset continue to be supported/hosted/maintained?}{If so, please describe how. If not, please describe how its obsolescence will be communicated to users.}

\dsanswer{As of the time of writing older versions are still supported. We could not find information related to what happens when a version is turned down, or when this would take place.}

\dsquestionex{If others want to extend/augment/build on/contribute to the dataset, is there a mechanism for them to do so?}{If so, please provide a description. Will these contributions be validated/verified? If so, please describe how. If not, why not? Is there a process for communicating/distributing these contributions to other users? If so, please provide a description.}

\dsanswer{Yes, community contributions are welcomed and are typically performed through GitHub. More information can be found on \url{https://mimic.mit.edu/docs/community/contributing/}.}

\dsquestion{Any other comments?}

\dsanswer{
}

\end{multicols}
\end{singlespace}

\appendix
\newpage
\section*{Appendix B: Semi-structured expert interview}
We recruited experts with wide range of expertise using health data in relation to their job. Expertise are listed in Table 1.
During the interview, study objectives and research questions were presented. Interviews time varied between 30 and 180 minutes over one to two sessions, depending on the experts' availability and engagement. In certain cases, interviewees chose to offer additional comments after the interviews. 

We first asked questions pertaining to the participants expertise in health data and professional background. We then conducted a semi-structured interview using the protocol described bellow. Finally, participants and interviewers together studied the \textit{Primary Healthsheet}, discussing\footnote{The interview transcripts and notes are over 300 pages, providing comprehensive insights into a variety of aspects to consider when evolving future versions of the Healthsheet.} specific questions and potential opportunities for improvement. 

The following questions were asked from the expert interview participants, and then followed by a deep dive in questionnaire:
\begin{itemize}
\item Do you use healthcare data in relation to your job? If yes, which kinds of healthcare applications have you worked with?
\item What are the challenges you have faced when using or choosing a healthcare dataset? 
\item  If you have access to the healthsheets of all datasets, would you use it to decide which dataset to work with and if yes, how do you use Healthsheets to help you decide if you want to use a dataset or not? 

\item Among the given categories/ questions in the Primary Healthsheet, are there any categories or questions that you would prioritize having access to? 
\item Which questions are the most important questions for you and your assessment of a dataset?
\item If I am a curator of a dataset, what should be my incentive for creating something like this? Is it worth it to spend time and effort on filling this long questionnaire?
\item  It's a very large questioner, how can we make healthsheet more personalized for your role?

\end{itemize}

\begin{table}
\label{expert-table}
\begin{tabular}{c|c}
Participants Expertise               & Number of participants \\ \hline
Legal and Regulatory                 & 1                      \\
Clinical                             & 7                      \\
ML for Health Research & 6                      \\
Health Data Infrastructure & 2
\\
Product                              & 4                      \\
Bioethics                            & 1                      \\
Equity                               & 2                      \\
Privacy                              & 2                      \\
Sociotechnical Research              & 1    
    \\ \hline
Total number of participants & 21
\end{tabular}
\caption{Expertise of the interview participants. There were in total 21 participants. Some participants expertise are beyond one category and is on the intersection of 2 or more expertise.}
\end{table}

\end{document}